\documentclass[journal]{IEEEtran}
\usepackage{cite}

\usepackage{algpseudocode}
\usepackage{array}
\usepackage{enumitem}
 \usepackage{mathrsfs}
\usepackage{multirow}
\usepackage{footnote}
\usepackage{amsmath}
\usepackage{algorithm}
\usepackage{algorithmicx}
\usepackage[pdftex]{graphicx}
\newcommand{\ie}{\textit{i.e.}}

\newcommand{\etal}{\textit{et al.}}

\graphicspath{{pdf/}}
\DeclareGraphicsExtensions{.pdf}

\begin{document}

\title{Person Re-Identification via Active Hard Sample Mining}

\author{Xin~Xu,~\IEEEmembership{Member,~IEEE,}
       Lei~Liu,~\IEEEmembership{Student Member,~IEEE,}
       Weifeng~Liu,~\IEEEmembership{Senior Member,~IEEE,}
       Meng~Wang,~\IEEEmembership{Senior Member,~IEEE,}
       and~Ruimin~Hu,~\IEEEmembership{Senior Member,~IEEE}

\thanks{Xin Xu and Lei Liu are with the School of Computer Science and Technology, Wuhan University of Science and Technology, Wuhan 430081, China. Weifeng Liu is with the College of Control Science and Engineering, China University of Petroleum (East China). Meng Wang is with the School of Computer Science and Information Engineering, Hefei University of Technology, Hefei. Ruimin Hu is with the School of Computer Science, Wuhan University.
}
}

\markboth{}%
{Shell \MakeLowercase{\textit{et al.}}: Bare Demo of IEEEtran.cls for IEEE Journals}

\maketitle


\begin{abstract}
Annotating a large-scale image dataset is very tedious, yet necessary for training person re-identification models. To alleviate such a problem, we present an active hard sample mining framework via training an effective re-ID model with the least labeling efforts. Considering that hard samples can provide informative patterns, we first formulate an uncertainty estimation to actively select hard samples to iteratively train a re-ID model from scratch. Then, intra-diversity estimation is designed to reduce the redundant hard samples by maximizing their diversity. Moreover, we propose a computer-assisted identity recommendation module embedded in active hard sample mining framework to help human annotators to rapidly and accurately label the selected samples. Extensive experiments were carried out to demonstrate the effectiveness of our method on several public datasets. Experimental results indicate that our method can reduce 57\%, 63\%, and 49\% annotation efforts on the Market1501, MSMT17, and CUHK03, respectively, while maximizing the performance of the re-ID model.
\end{abstract}

\begin{IEEEkeywords}
Active Learning, Person Re-Identification, Hard Sample Mining.
\end{IEEEkeywords}

\IEEEpeerreviewmaketitle

\section{Introduction}
\IEEEPARstart{P}{erson} re-identification (re-ID) aims to match a specific pedestrian across different cameras. This is an essential task for public security. Several efforts have been dedicated to the person re-ID problem. Starting from classical LOMO~\cite{liao2015person} and BoW~\cite{zheng2015scalable}, recent state-of-the-art approaches have turned to convolutional networks~\cite{chang2018multi,song2018mask,zheng2019pose} and observed a further performance improvement. Specifically, these methods utilize discriminative embedding features to better overcome typical challenges, including appearance, illumination, and occlusion.

\begin{figure}[!t]
	\setlength{\abovecaptionskip}{0.cm}
	\setlength{\belowcaptionskip}{-0.cm}
	\centering
	\includegraphics[width=1\columnwidth]{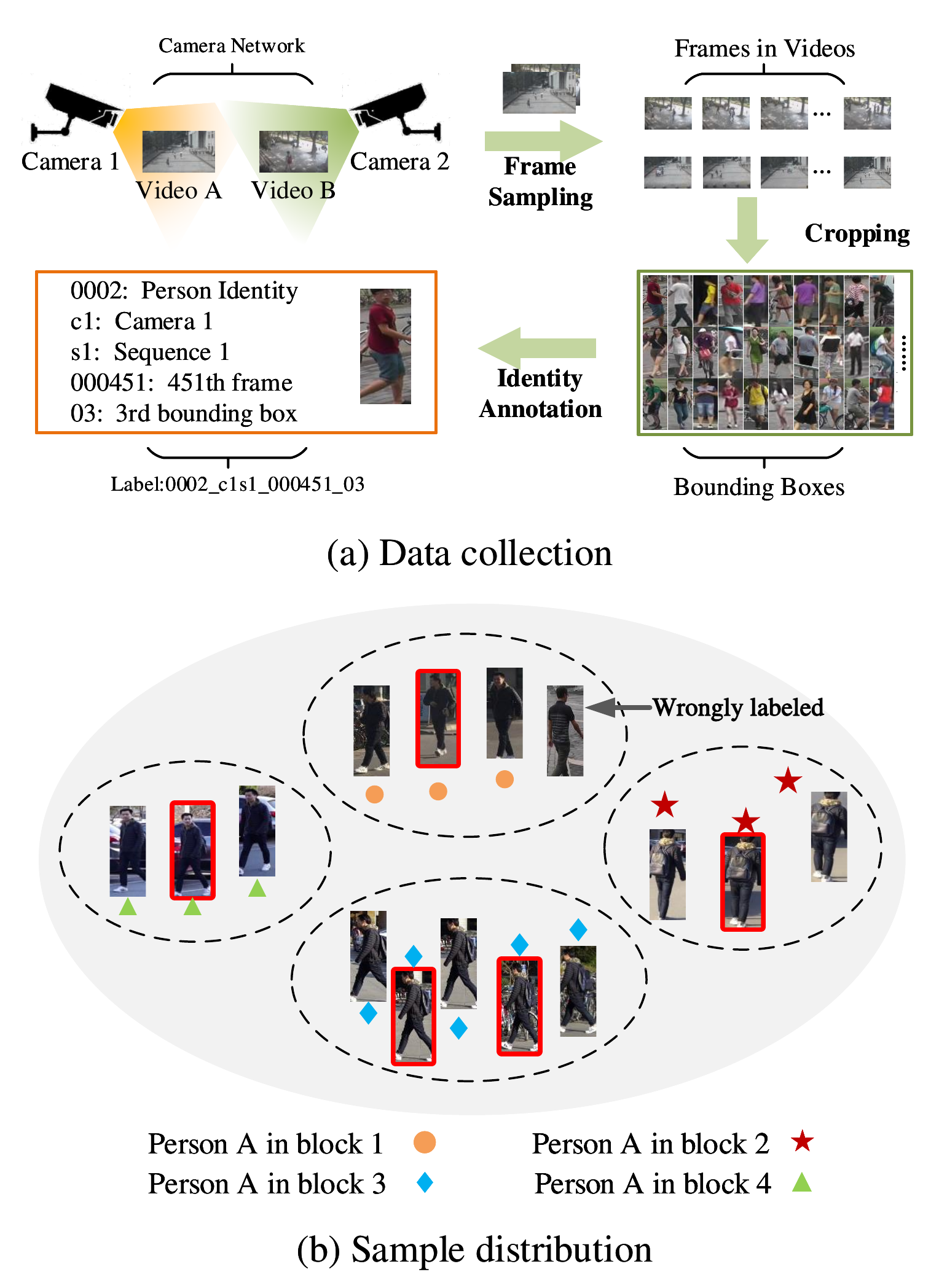}
	\caption{Illustration of data collection and sample distribution in person re-ID. (a) The data collection consists of video collection, frame sampling, bounding box cropping, and identity annotation. And (b) samples of person A are clustered into different blocks with redundancies due to high intra-class variance. There may exist redundancy in these blocks. The images with red boxes are effective samples of different blocks.}
	\label{fig1}
\end{figure}

Most existing person re-ID methods are supervised learning. They hence require large amounts of labeled data for training, which is labor- and time-consuming. Figure \ref{fig1}(a) demonstrates the detailed procedure of data collection with expensive human efforts, containing four steps: video collection, frame sampling, bounding box cropping, and identity annotation.

Numerous approaches have been proposed to address the data collection problem in person re-ID. Weakly supervised learning attempts to learn the discriminative information from weak annotations~\cite{meng2019weakly} or partially labeled dataset~\cite{zhang2016learning,zhu2017semi}. Other investigations utilize unsupervised settings~\cite{yu2017cross,wang2018transferable,yang2017unsupervised} or transfer learning~\cite{shi2015transferring,liu2018pose} to employ unlabeled data. However, the performance of these methods is generally inferior~\cite{kodirov2015dictionary,wang2018transferable}. The main reason is the lack of labeled cross-view discriminative information in every camera pair, since the amount of labeled data is a significant factor influencing the performance of a person re-ID system.

Active learning is a natural way to address data collection problem, and has been used to reduce annotation efforts for several computer vision tasks, such as human pose estimation~\cite{liu2017active}, medical image processing~\cite{hoi2006batch}, image classification~\cite{mccallumzy1998employing}, and semantic segmentation~\cite{vezhnevets2012active,konyushkova2015introducing}. However, the data collection problem remains largely untapped for person re-ID, due to the high intra-class variance of the same person and low inter-class variance of different persons. As illustrated in Figure \ref{fig1}(b), images of one person are classified into different blocks due to high intra-class variance. What's more, there exist wrongly labeled images during data collection in real scenarios.

To train an effective person re-ID model with the least labeling efforts, we focus on learning from scratch with incremental labeling via human annotators and model feedback. This method differs from the train-once-and-deploy scheme~\cite{liu2019deep}, which annotates all person images before training a re-ID model. Therefore, an incremental annotation process of active learning is adopted to select informative samples from an unlabeled set in each iteration. Then these samples are labeled by human annotators to update the model. Considering that hard samples can provide informative patterns, above sample selection and model updating process naturally incorporates hard sample mining and active learning together.

In this paper, we present an Active Hard Sample Mining (AHSM for short) framework to address the data collection problem in person re-ID, which aims to train an effective re-ID model with the least labeling efforts. Existing active learning based methods only focus on selecting informative samples, but ignore the redundancy among them. The proposed AHSM can automatically select hard samples containing informative patterns, and reduce redundant hard samples simultaneously. Hard samples generally are low confidence samples, and can be measured by uncertainty estimation. Redundant hard samples are reduced via intra-diversity estimation to maximize the diversity of hard samples of a person. After being labeled by human annotators, the selected hard samples are progressively fed into the training set to retrain the re-ID model until the desired performance. We conduct extensive experiments on three public person re-ID datasets. Experimental results indicate that our method can reduce 57\%, 63\%, and 49\% annotation efforts on the Market1501, MSMT17, and CUHK03, respectively, while maximizing the performance of the re-ID model.

The main contribution of our work is threefold:
\begin{itemize}
	\item To address the data collection problem in person re-ID, we present a novel active learning framework incorporating hard sample mining and redundancy reduction to alleviate the labeling efforts.
	\item To minimize annotation efforts while maximizing the performance of the re-ID model, we design an uncertainty estimation to select hard samples with informative patterns. Then, intra-diversity estimation is formulated to remove redundant hard samples.
	\item A computer-assisted Identity Recommendation Module (IDRM) is proposed to help the human annotators to rapidly and accurately label the selected hard samples. Besides, IDRM can significantly reduce wrong annotations in public person re-ID datasets.
\end{itemize}

The rest of this paper is organized as follows. Section II reviews the related studies of person re-ID and active learning. In Section III, we detail the general framework and the proposed estimation methods. Section IV analyzes the experimental results and the ablation studies. Section V summarizes the paper and presents the future work.

\section{Related Work}
Numerous deep models have achieved great successes in person re-ID. Yet the main challenge is the limited amount of labeled samples with expensive labeling cost. Regarding the data collection in re-ID, the scope of collected dataset is relatively limited and partial compared to the spatial and temporal distribution of real data. At the same time, the scale of datasets in re-ID is very small compared with other computer vision tasks. According to the manner of data utilization, re-ID approaches can be divided into three groups: learning from full annotations, learning from weak annotations, and active learning. In the following subsections, we detail them one by one.

\subsection{Learning from full annotations}
Person re-ID can be viewed as an image retrieval task~\cite{zheng2014coupled,zheng2017sift,zheng2014mathcal}. It aims to match a probe person of interest across multi-camera views against a series of gallery images. In the full annotations setting, most existing supervised person re-ID methods~\cite{bai2017scalable} have progressed on several large-scale datasets in recent years. The studies~\cite{tian2018eliminating,li2018harmonious,varior2016learning,lisanti2014person,cheng2016person,li2017learning} are proposed to learn discriminative and powerful representations for person description. For example, Su $\etal$~\cite{su2017pose} proposed a pose-driven deep re-ID model to learn robust feature representations with pose variations. Similarly, Wei $\etal$~\cite{wei2017glad} extracted global-local-alignment descriptors via human pose estimation. In~\cite{xiao2016learning}, data from several domains is utilized to train an effective model to extract discriminative features. Besides feature extraction, many re-ID methods focus on distance metric learning~\cite{bak2017one,zheng2015partial,ma2014person,zhu2017fast} and end-to-end deep learning~\cite{ahmed2015improved,xiao2016learning,zhao2017deeply,liu2017end}. These efforts generally rely on abundant labeled data. However, the data collection is particularly tedious and typically involves thousands of hours of human efforts. Therefore, this paper focuses on reducing the labeling efforts in the fully-supervised setting.

\subsection{Learning from weak annotations}
One theme in reducing annotation efforts for person re-ID is to learn from weak annotations. In~\cite{meng2019weakly}, person re-ID model learns from the annotation only containing the name of the identity. Another solution is based on the semi-supervised learning setting~\cite{liu2018semi,li2018semi,zhang2016learning}, whereby the annotations for some images are avoided. For example, Liu $\etal$~\cite{liu2014semi} jointly learned from two coupled dictionaries for gallery and query cameras (both labeled and unlabeled images in the training process) to explore the person appearance variations across cameras. In~\cite{fan2018unsupervised}, pre-trained person representations are transferred to unseen domains with a few labels. Moreover, some approaches learn identity information in an unsupervised setting where the annotation is not necessary~\cite{fan2018unsupervised,yu2018unsupervised,peng2016unsupervised}. For example, Deng $\etal$~\cite{deng2018image} transferred the labeled identity information from source domain to target domain with the constraints of self-similarity and domain-dissimilarity. In~\cite{wang2018transferable}, an attribute-semantic and discriminative feature is learnt and transferred to target domains for person re-ID task.

However, all of these efforts assume the labeled data is given and fixed. Besides, these methods generally exhibit performance degradation compared to the fully-supervised methods. Therefore, we focus on the fully-supervised active learning setting whose goal is to estimate which person image worths labeling. Once an image is selected, full annotations are given by human annotators.

\subsection{Active learning}
Active learning is a momentous sub-domain of machine learning, with plenty of query heuristics~\cite{wang2016cost,gal2017deep,beluch2018power} proposed for measuring the effectiveness of an as-yet unlabeled sample. General query heuristics include entropy~\cite{lewis1994heterogeneous}, reducing the expected error of the classifier~\cite{vijayanarasimhan2010visual}, maximizing the diversity among the selected samples~\cite{hoi2009semisupervised}, or maximizing the expected labeling change~\cite{vezhnevets2012weakly}.
In computer vision, active learning has been widely developed for addressing the data collection problem in various tasks such as classification~\cite{li2014multi}, recognition~\cite{lin2017active}, and object detection~\cite{wang2018towards}. However, quite a few attempts have utilized the active learning to solve the data collection problem for person re-ID thus far. In~\cite{liu2019deep}, data uncertainty is utilized as the objective function of the reinforcement learning policy, and feature similarity is used to construct the reinforcement learning state. However, it ignores the redundant samples and feature effectiveness during the model training. Therefore, we present an active learning framework to minimize annotation efforts while maximizing the performance of the person re-ID model, where redundant samples would be left out.

\begin{figure*}[!t]	
	\setlength{\abovecaptionskip}{0.cm}
	\setlength{\belowcaptionskip}{-0.cm}
	\centering
	\includegraphics[width=1\textwidth]{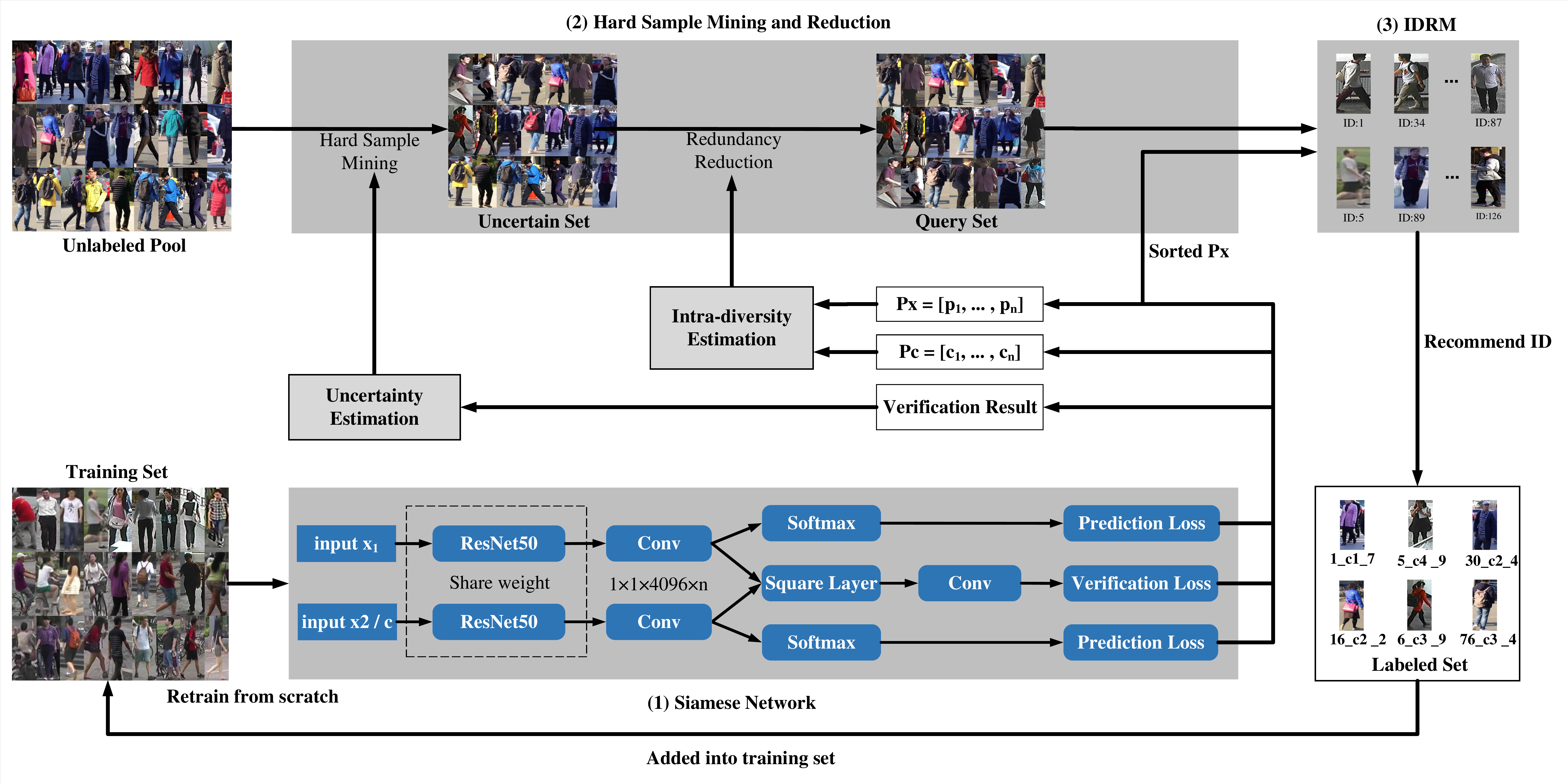}
	\caption{Schematic illustration of our proposed AHSM framework for person re-ID. (1) Siamese Network combines identification loss and verification loss. (2) Uncertainty and intra-diversity estimation are used to select hard sample and reduce redundancy. And (3) IDRM is a computer-assisted interface to facilitate human annotators. We begin with a randomly initialized Siamese Network. Following that, a hard sample mining method searches the large unlabeled image pool for hard samples/images without redundancy. IDRM further saves the search time by recommending a few candidate images to human annotators.}
	\label{fig2}
\end{figure*}

\section{Active Learning for Person Re-ID}
The framework of AHSM consists of four sub-modules, i.e., Siamese Network (SN), uncertainty estimation, intra-diversity estimation, and IDRM. SN combines identification loss and verification loss. Uncertainty and intra-diversity estimation are used to select hard sample and reduce redundancy. IDRM is a computer-assisted interface to facilitate human annotators. Figure \ref{fig2} shows the framework of AHSM. We detail the AHSM framework and its sub-modules as follows.

\subsection{General Framework}
We have an unlabeled re-ID dataset $\mathcal{U}=\left\{x_{i}\right\}^{n}$, where $x_{i}$ is an unlabeled person image collected in video surveillance and $n$ is the size of $\mathcal{U}$. Our goal is to minimize annotation efforts while maximizing the performance of the person re-ID model.

The general framework of AHSM is illustrated in Algorithm \ref{alg:Framework} and Figure \ref{fig2}. The framework of AHSM consists of four sub-modules: a siamese network (SN)~\cite{zheng2018discriminatively}, uncertainty estimation, intra-diversity estimation, and IDRM. SN is randomly initialized at first. Then, uncertainty estimation is proposed to mine hard samples. And intra-diversity estimation is employed to reduce the redundant hard samples. The remained hard samples are moved into $\mathcal{Q}=\left\{q_{i}\right\}^{t}$. A computer-assisted interface IDRM is introduced to help the human annotators to rapidly and accurately label the subset $\mathcal{Q}$. After being labeled, the subset $\mathcal{Q}$ is added into the training set to update SN. Sample selection and training are iterated until the model delivers the required accuracy.

\begin{algorithm}[!t]  		
	\caption{Algorithm of our proposed AHSM}
	\label{alg:Framework}
	\begin{algorithmic}[1]
		\Require ~~\\
		$U:$ The unlabeled dataset;\\
		$T:$ The training set;\\
		$Q:$ The query set;
		\Ensure ~~\\
		Initial SN;
		\Repeat
		\For{each $x \in \mathcal{U}$}
		\State calculate uncertainty value of $x$;
		\State sort $x$ in descending order according to their uncertainty values;
		\EndFor
		\State select hard samples according to sorted list of $x$;
		\For{each $x_{h} \in $hard samples}
		\State calculate intra-diversity value of $x$;
		\State sort $x_{h}$ in descending order according to their intra-diversity values;
		\EndFor
		\State remove redundant hard samples according to sorted list of $x_{h}$;
		\State select remained hard samples into $Q$;
		\State label $Q$ by human annotators and IDRM;
		\State add $Q$ into $T$;
		\State train SN based on $T$ from scratch;	       		       		
		\Until{query budget or expected performance is reached}
	\end{algorithmic}
\end{algorithm}

\subsection{Siamese Network}
Following~\cite{zheng2018discriminatively}, we formulate the person re-ID task as a verification task and a classification task simultaneously. Accordingly, we use a siamese network as the basic model for active learning. As shown in Figure \ref{fig2}, SN consists of two ResNet50 models, three additional convolution Layers, one Square Layer, and three losses. The weights of ResNet50 models are shared like a siamese structure.

During training, the identification labels carry the relations between person images and person identities, and the identification loss is calculated as following:
\begin{equation}
Loss_{identification}=\sum_{i=1}^{K}-p_{i} \log \left(\hat{y}_{i}\right),
\end{equation}
where $\hat{y}_{i}$ is the predicted probability and ${p}_{i}$ is the target probability (0 or 1). The verification labels carry the relations among person images. And the verification loss is calculated as:
\begin{equation}
Loss_{verification}=\sum_{i=1}^{2}-q_{i} \log \left(\hat{q}_{i}\right),
\end{equation}
where $\hat{q}_{i}$ is the predicted probability and $\hat{q}_{1}+\hat{q}_{2}=1$. If the image pair have the same identity, $q_1 = 1$, $q_2 = 0$; otherwise, $q_1 = 0$, $q_2 = 1$.

Note that both classification and verification compute the cross entropy between the prediction and true label. Thus, the main objective function can be formulated as following:
\begin{equation}
\mathrm{Loss}=-\sum_{\mathrm{i}=1}^{\mathrm{n}} \mathrm{p}_{\mathrm{i}} \log \left(\mathrm{y}_{\mathrm{i}}\right),
\label{for3}
\end{equation}
with
\begin{equation}
\begin{aligned}
\mathrm{y}_{\mathrm{i}}&=softmax(\mathrm{a}_{\mathrm{i}})\\
 &=\frac{e^{\mathrm{a}_{i}}}{\sum_{j=1}^{n} e^{\mathrm{a}_{j}}},
\label{for4}
\end{aligned}
\end{equation}
where $\textit{y}_{\mathrm{i}}$ is the predicted probability of a sample belonging to class $i$. $\textit{p}_{\mathrm{i}}$ is the one-hot encoding label of a sample. $N$ is the number of classes and $\textit{a}$ is the outputs of SN.

\subsection{Hard Sample Mining for Active Learning}
Hard sample mining can accelerate the convergence of the re-ID model. To minimize annotation efforts while maximizing the performance, the key is to make model convergence as quick as possible. Thus, we should select the samples with more contributions to improve the performance.
According to Equation~(\ref{for4}), we can get $\textit{d} y_{\mathrm{i}}/\textit{d} a$ as follows:

\begin{equation}
\frac{\textit{d} y_{\mathrm{i}}}{\textit{d} a_{\mathrm{i}}}=\left\{\begin{array}{ll}{\mathrm{y}_{\mathrm{i}}\left(1-\mathrm{y}_{\mathrm{i}}\right)} & {\text { if } i = j} \\
{-\mathrm{y}_{\mathrm{j}} \mathrm{y}_{\mathrm{i}}} & {\text { if } i \neq j}\end{array}\right..
\label{for5}
\end{equation}

\noindent Likewise, we can get $\textit{d} Loss/\textit{d} y_{\mathrm{i}}$ according to Equation~(\ref{for3}) as follows:

\begin{equation}
\frac{\textit{d} Loss}{\textit{d} y_{\mathrm{i}}}=-\mathrm{p}_{\mathrm{i}} \frac{1}{\mathrm{y}_{\mathrm{i}}}.
\label{for6}
\end{equation}

\noindent According to Equation~\ref{for5} and ~\ref{for6}, $\textit{d} Loss/\textit{d} a_{\mathrm{i}}$ can be calculated as following:
\begin{equation}
\vert\frac{\textit{d} Loss}{\textit{d} a_{\mathrm{i}}}\vert=\vert\mathrm{y}_{\mathrm{j}}-\mathrm{p}_{\mathrm{j}}\vert.
\end{equation}

In order to make model convergence as quick as possible, we expect a large value of $\vert\textit{d} Loss/\textit{d} a_{\mathrm{i}}\vert$, only larger when the samples contain more uncertain identity information with classification errors. These samples are defined as hard samples for active learning. Thus, we propose an uncertainty estimation for hard sample mining.

\subsubsection{\textbf{Uncertainty Estimation for Hard Sample Mining}}
Uncertainty has been widely utilized for sample selection in active learning~\cite{yang2015multi,yan2016image}. In this work, we also define the uncertainty of the samples but in a novel calculation way. Combining identification branch with verification branch of SN, the uncertainty of a sample is defined as follows:

\begin{equation}
Uncertainty(x)=1-\operatorname{Verification}\left(S_{ID(x)}, x\right),
\end{equation}

\noindent where $x$ is an unlabeled person image, and $\operatorname{Verification}()$ is the verification branch. $S_{ID(x)}$ is a sample in the labeled set that has the same ID with $x$, and the ID of $x$ can be predicted as:
\begin{equation}
ID(x)=\operatorname{Prediction}(\mathrm{x}),
\end{equation}
\noindent where $\operatorname{Prediction}()$ is the identification branch. To get precise uncertainty, we use the average of the class to replace $S_{ID(x)}$. If two branches gave contradictory results for one sample, this sample is defined as a hard sample.

\begin{figure*}[!t]
	\setlength{\abovecaptionskip}{0.cm}
	\setlength{\belowcaptionskip}{-0.cm}
	\centering
	\includegraphics[width=1\textwidth]{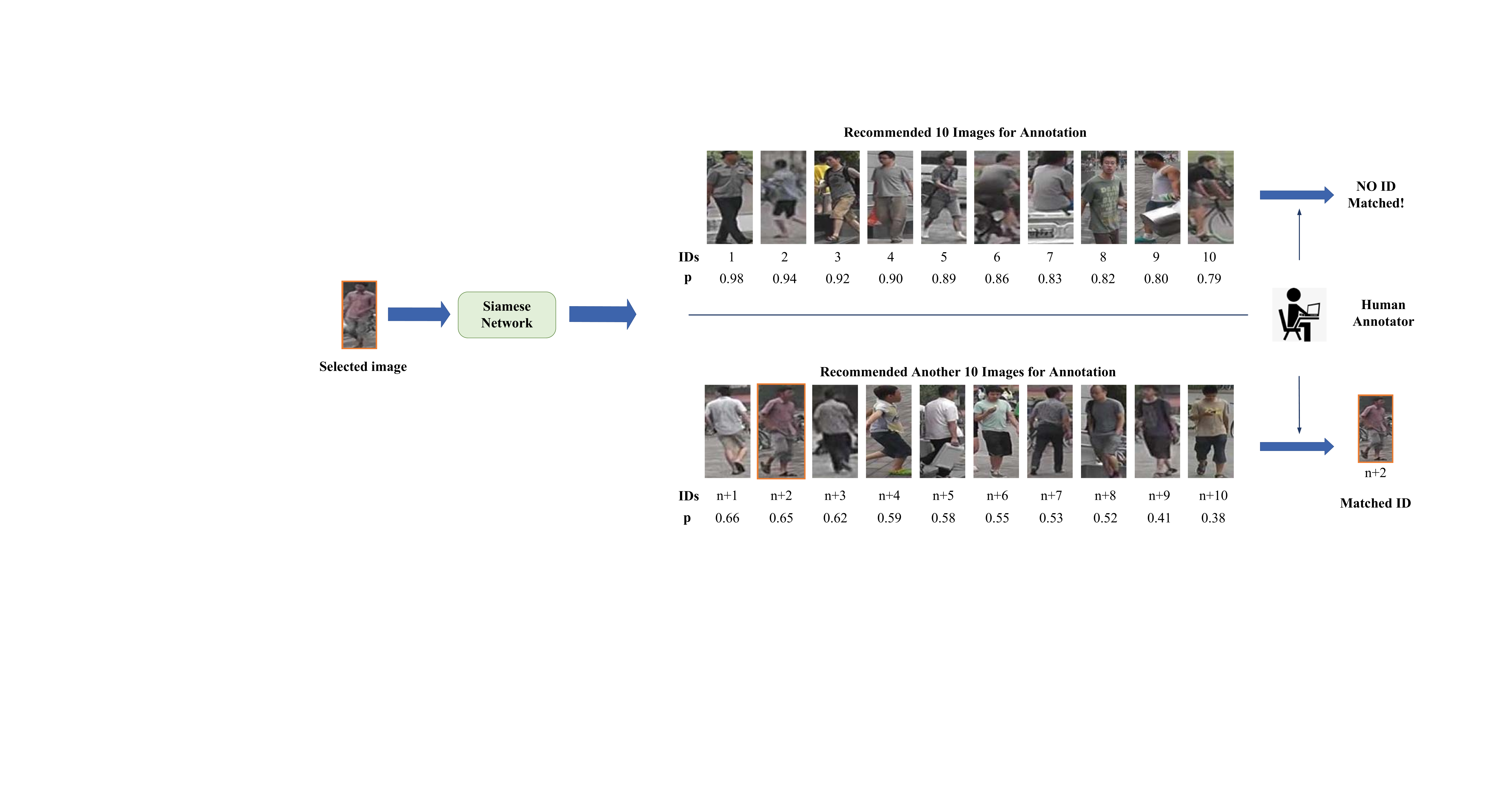}
	\caption{Illustration of the computer-assisted interface IDRM. (1) IDRM can recommend candidate images for an unlabeled image according to the predicted probabilities obtained from the reID models. (2) Human annotators match the unlabeled image and candidated images for annotation. The recommendation is iterated until the image is labeled successfully. With the cooperation of IDRM and human annotators, wrong annotations and labeling efforts can be significantly reduced.}
	\label{fig3}
\end{figure*}

\subsubsection{\textbf{Intra-diversity Estimation for Redundancy Reduction}}
Intra-diversity estimation is designed to reduce the redundant hard samples by maximizing the diversity of intra-class hard samples. High intra-class variance may cause redundant samples of the same person. Thus it exists redundancy among the hard samples selected by uncertainty estimation. To reduce the redundancy and maintain the performance, the training set should reduce hard yet similar samples but keep the comprehensive identity information for each identity. To this end, the main idea is to select different hard samples of a person by the similarity measurement between an unlabeled hard sample and the labeled samples.

Similarity can be well measured by the feature distance in the Euclidean space. However, the training subset is updated incrementally, and may be not sufficient to extract effective features at current stage. Besides, we expect the large diversity of hard samples in training set for each identity. Therefore, we introduce the Kullback Leibler (KL) divergence to calculate the difference between an unlabeled sample and the labeled samples. The KL divergence can be calculated as:

\begin{equation}
Distance(P, Q)=\sum_{k=1}^{K}\left(p_{i}-q_{i}\right) \log \frac{p_{i}}{q_{i}}.
\end{equation}

\noindent There are two advantages of KL divergence: (I) Using KL divergence can select different hard samples with no effect from the feature effectiveness, since the predicted probability distributions of two samples can reveal their similarity. And (II) different predicted probability distributions can improve the diversity of hard samples in training set via KL divergence. Then we introduce the detailed process of redundancy reduction via intra-diversity estimation.

Given hard samples selected by uncertainty estimation, the feature of each sample is extracted by the re-ID model. For an accurate intra-diversity, we take the center of the class to replace one labeled sample in the calculation. Denote the labeled set by $L$, then the center of each class in $L$ can be calculated by:

\begin{equation}
Center=\frac{1}{n} \sum_{i}^{n} l_{i}^{k},
\end{equation}

\noindent where $l_{i}^{k}$ is the feature of $i\mbox{-}th$ sample in $k\mbox{-}th$ class. Then the intra-diversity can be formulated as:

\begin{equation}
Intra\mbox{-}diversity=\sum_{i}^{n}\left(p_{i}-c_{i}\right) \log \frac{p_{i}}{c_{i}},
\end{equation}

\noindent where $p_{i}$ represents the predicted probability of the currently unlabeled sample belonging to the $i\mbox{-}th$ person class. $c_{i}$ represents the predicted probability of the class center in $L$ from the $i\mbox{-}th$ person class. The maximization principle can guarantee low redundancy and comprehensive identity information. Therefore, we query samples by intra-diversity according to the maximization principle as follows:

\begin{equation}
S_{query}=\underset{x}{\operatorname{argmax}}(Intra\mbox{-}diversity(x)),
\end{equation}

\noindent where $x$ represents a sample in the uncertain set. In this case, we can reduce the labeling efforts by reducing redundant hard samples.

Besides, we select samples in sample space only relying on intra-diversity value. Moreover, at each iteration, the labeled set and SN model can be updated incrementally. The update strategy can guarantee that the distribution of the labeled set can approximate the real sample distribution.

\subsection{IDRM for Labeling Efforts Reduction}
We propose a computer-assisted interface IDRM to help human annotators to rapidly and accurately label the selected samples. In the conventional setting, human annotators must compare images one-by-one to annotate the unlabeled images. As shown in Figure \ref{fig3}, IDRM can recommend candidate images to human annotators for an unlabeled image, which enables human annotators to label images based on the candidates to largely reduce the labor burden.

In addition, it is observed that there are many wrongly labeled images in public datasets. To reduce the wrong annotations and obtain the effective recommendations, we sort the candidate images according to the predicted probabilities obtained from the SN model. The high probability of a sample indicates the high matching possibility between this sample and the given identity. Top 10 candidates are recommended to help human annotators to rapidly and accurately label the selected images. If the match fails, another 10 candidates are provided according to the sorted list until all selected images are labeled. With the cooperation of IDRM and human annotators, wrong annotations can be significantly reduced.

\section{Experiments}
We conducted extensive comparison experiments and ablation studies on three public re-ID datasets to validate the effectiveness of our proposed AHSM framework. Specifically, the three person re-ID datasets and the evaluation protocols are firstly introduced. Then, we detail the experimental settings, random selection setting, and query heuristics for comparison. Afterward, the performance comparison and the ablation studies of AHSM are given. Finally, we present experimental analyses of AHSM.

\begin{figure}[!t]
	\setlength{\abovecaptionskip}{0.cm}
	\setlength{\belowcaptionskip}{-0.cm}
	\centering
	\includegraphics[width=1\columnwidth]{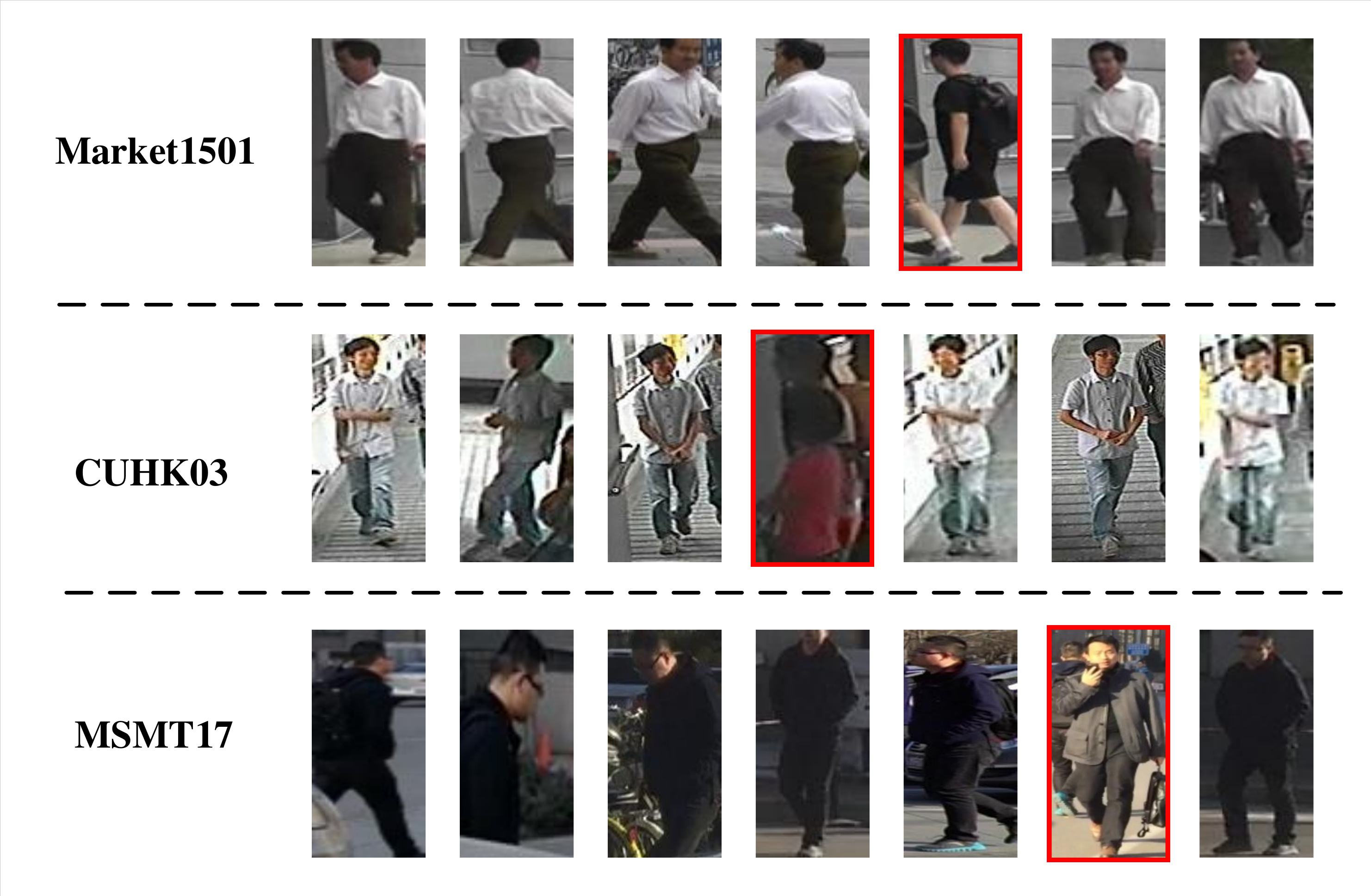} 
	\caption{Sample images from three public person re-ID datasets. The images with red boxes are wrongly labeled samples of the three datasets.}
	\label{fig4}
\end{figure}

\begin{table}[!t]
	\setlength{\abovecaptionskip}{0.cm}
	\setlength{\belowcaptionskip}{-0.cm}
	\caption{The characteristics of three person re-ID datasets.}
	\label{datasets}
	\centering
	\renewcommand\arraystretch{1.5}
	\fontsize{8}{9}\selectfont
	\setlength{\tabcolsep}{5mm}{
		\begin{tabular}{cccc}
			\hline
			\textbf{Dataset} &\textbf{\# ID} &\textbf{\# BBoxes} &\textbf{\# Cam} \\ \hline
			Market1501 ~\cite{zheng2015scalable} & 1501 &32668&6 \\
			MSMT17~\cite{wei2018person}  &  4101  &126441&15\\
			CUHK03~\cite{li2014deepreid} &  1467  &14097&5(pairs)\\
			\hline
	\end{tabular}}
\end{table}

\subsection{Datasets}
We conducted extensive experiments on the Market1501~\cite{zheng2015scalable}, MSMT17~\cite{wei2018person}, and CUHK03~\cite{li2014deepreid}. These datasets are widely used in the person re-ID field, among which the MSMT17 is the largest public person re-ID dataset. Table \ref{datasets} shows a statistical review of the three datasets. We indicated the number of person identities (ID), bounding boxes (BBoxes), and total cameras (Cam) in each dataset. Figure \ref{fig4} shows some image samples and wrongly labeled samples from the three datasets.
\subsubsection{\textbf{Market1501}}
The Market1501 dataset has 32,668 labeled bounding boxes of 1,501 identities. In detail, the training set has 12,936 bounding boxes of 750 identities, while the testing set contains 19,732 bounding boxes of 751 identities. In the testing set, one image of each identity is randomly selected as probe image for each camera. The testing set totally has 3,368 probe images. And there are 2,793 bounding boxes as distractors in the gallery set. Besides, a total of 6 cameras are used. The person boxes are cropped from the raw video frames by the DPM detector. Each person identity has more than one image captured by at most six cameras and at least two cameras. The standard protocol~\cite{zheng2015scalable} is adopted for Market1501.
\subsubsection{\textbf{MSMT17}}
The MSMT17 dataset is collected in a campus. And it has similar viewpoint with Market1501, but more complicated scenarios. The MSMT17 dataset contains 126,441 labeled bounding boxes of 4,101 identities. It has 1,041 training identities and 3,060 testing identities. The testing set totally has 11,659 probe images and 82,161 gallery images. Totally 15 cameras are used including 12 outdoor cameras and 3 indoor cameras. The images are captured in 4 days with different weather within a month. 3 one-hour videos are selected from morning, noon and afternoon, respectively. The person images are cropped from original video frames by Faster RCNN detector. This dataset is the largest person re-ID dataset so far~\cite{wei2018person}.
\subsubsection{\textbf{CUHK03}}
The CUHK03 dataset is the first person re-ID dataset that is large enough for deep learning~\cite{li2014deepreid}. The CUHK03 dataset has 14,097 labeled bounding boxes of 1,467 identities. And it has 100 identities for training and 1,160 identities for testing. Totally 5 pairs of cameras are used. The person images are manually cropped and automatically cropped respectively. The detection model is deformable part models(DPM). The quality of person detection is relatively good for CUHK03.

\subsubsection{\textbf{Evaluation Protocols}}
The standard protocol is utilized on the three datasets for fair comparison experiments. The testing protocols of the three datasets are as follows. (I) For Market1501 dataset, it is randomly split into two parts. Half of the person identities are used for training, while the rest are used for testing. (II) For MSMT17 dataset, it is randomly split into training set and testing set. The ratio of the training set to the testing set is 1:3. (III) For CUHK03 dataset, we followed the standard protocol~\cite{li2014deepreid}, repeated it 20 times to randomly split the dataset into 100 persons for testing and the rest persons for training. Rank-1, Rank-5, Rank-10 accuracy and mean average precision (mAP) are computed to evaluate the performance of the re-ID models on the three datasets.

\subsection{Experimental Settings}
\subsubsection{\textbf{Parameters Setting}}
In our experiments, ResNet50 is adopted as the pre-trained model. During training, mini-batch stochastic gradient descent is used to update the parameters of the model. The learning rate is initialized as 0.01. Other parameters can be referred to~\cite{zheng2018discriminatively}. The experiments are repeated 10 times with average results reported. At each AHSM iteration, we selected $5\%$ data as the query subset.

\subsubsection{\textbf{Random Selection Setting}}
In this work, we adopted random selection as basic comparison method. The random selection method employs full training set to train the re-ID model. For each iteration, a query set is randomly selected from the unlabeled set. And we selected $5\%$ data as the query subset. After labeled, they are added into the training set to update the re-ID model.

\subsubsection{\textbf{Comparison Heuristics Setting}}
The following three query heuristics are utilized for comparison, which are basic and commonly used in active learning. We compared AHSM against them for all experiments.

\textbf{EP} This heuristic~\cite{joshi2009multi} ranks all the unlabeled samples in an ascending order according to their entropy value. Thus, EP is defined as:
\begin{equation}
ep_{i}=-\sum_{j=1}^{m} p(y_{i}=j | x_{i}) \log p(y_{i}=j | x_{i}).
\end{equation}
This heuristic takes all predicted probabilities into considering to select the informative samples.

\textbf{LC} This heuristic~\cite{scheffer2001active} ranks all the unlabeled samples in an ascending order according to the LC value, which is defined as:
\begin{equation}
lc_{i}=\max _{j} p(y_{i}=j | x_{i}).
\end{equation}
If the probability of the most probable identity (class) for an unlabeled sample is low, the sample is selected for annotation.

\textbf{MS} This heuristic~\cite{shannon1948mathematical} ranks all the unlabeled samples in an ascending order according to the MS value, which is defined as:
\begin{equation}
ms_{i}=p(y_{i}=j_{1} | x_{i})-p(y_{i}=j_{2} | x_{i}),
\end{equation}
where $j_{1}$ and $j_{2}$ represent the first and second most probable identity predicted by the re-ID model, respectively. The samples are selected according to the margin between $j_{1}$ and $j_{2}$.

\begin{figure*}[!t]
	\centering
	\includegraphics[width=1\textwidth]{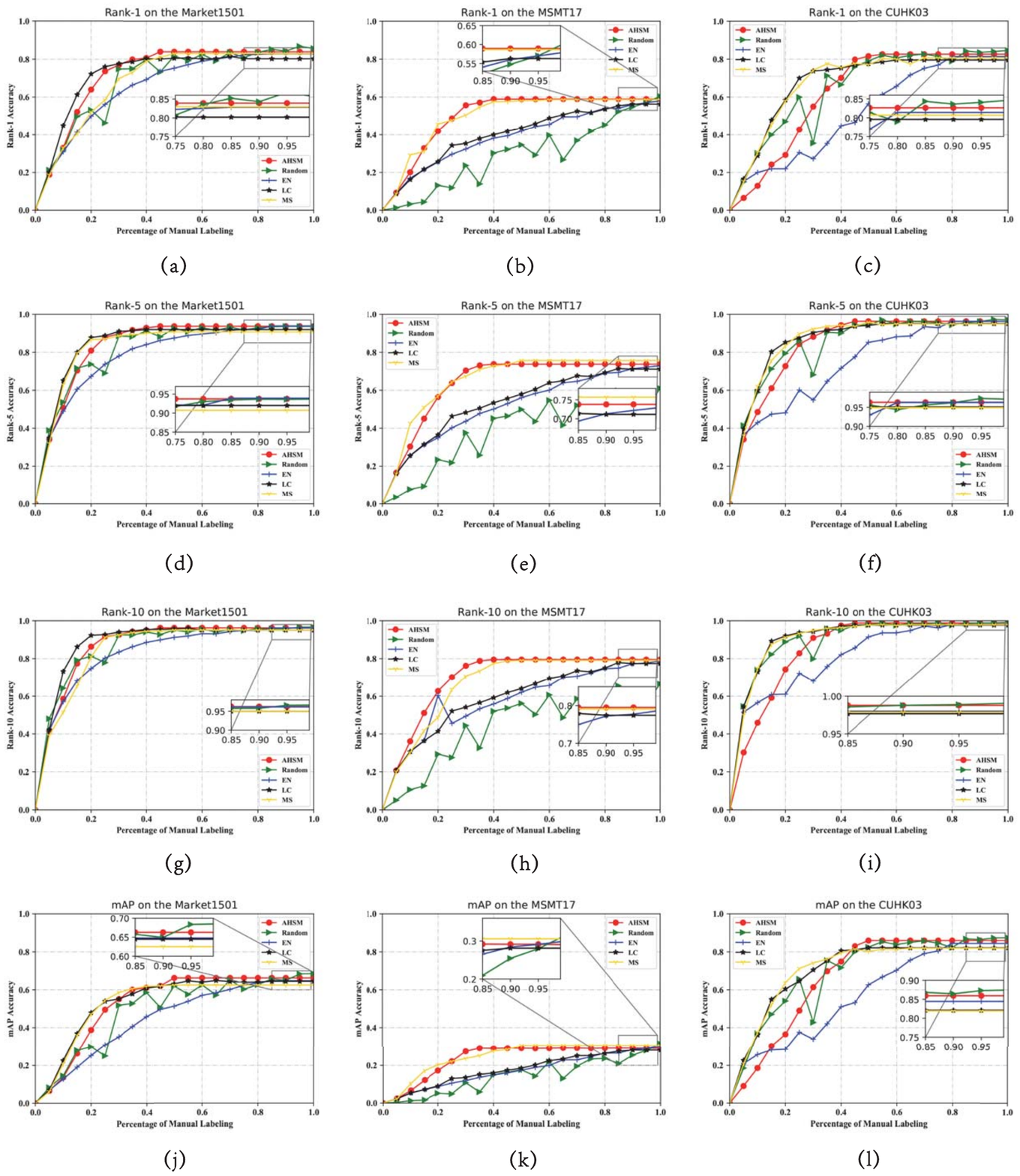} 
	\caption{Performance comparisons of the AHSM with random selection, entropy, LC and MS under different percentages of annotated samples on the three datasets. The red/rhombus curves are for AHSM, the green/triangle curves are for random selection, the blue/plus curves are for entropy, the yellow curves are for LC and the black/$*$ curves are for MS. (a)-(c) are for rank-1, (d)-(f) are for rank-5, (g)-(i) are for rank-10 and (j)-(l) are for mAP. (a, d, g, j) are curves on the Market1501 dataset, (b, e, h, k) are curves on the MSMT17 dataset, and (c, f, i, l) are curves on the CUHK03 dataset.}
	\label{fig5}
\end{figure*}

\begin{figure*}[!t]
	\centering
	\includegraphics[width=1\textwidth]{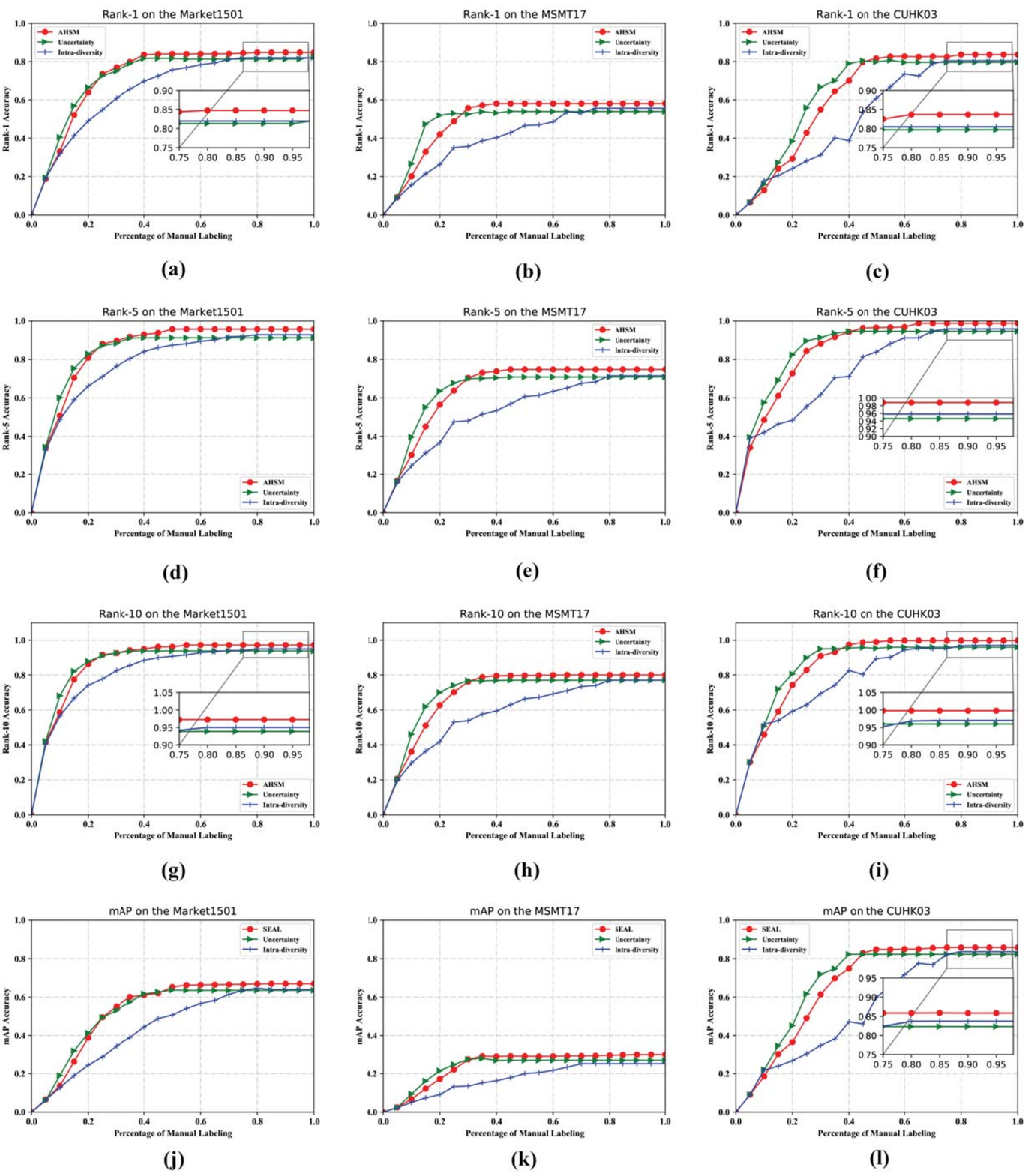} 
	\caption{Performance comparisons of the ablation studies under different percentages of annotated samples on the three datasets. The red/rhombus curves are for AHSM, green/triangle curves are for uncertainty heuristic, and blue/plus curves are for intra-diversity heuristic. (a)-(c) are for rank-1, (d)-(f) are for rank-5, (g)-(i) are for rank-10 and (j)-(l) are for mAP. (a, d, g, j) are curves on the Market1501 dataset, (b, e, h, k) are curves on the MSMT17 dataset, and (c, f, i, l) are curves on the CUHK03 dataset.}
	\label{fig6}
\end{figure*}

\subsection{Experimental Results}
\subsubsection{\textbf{Comparison with Random Selection}}
In Table \ref{market}, \ref{msmt}, and \ref{cuhk}, the performance of AHSM is compared with random selection on the three datasets using different metrics including rank-1, rank-5, rank-10, mAP, and percentages of annotated data. Moreover, we illustrated the performance comparisons of AHSM with random selection under different percentages of annotated samples on the three datasets in Figure \ref{for4}. In Table \ref{size}, we compared the training set sizes in detail on the three datasets.

\textbf{Results on the Market1501} In Table \ref{market}, we compared the performance of AHSM with random selection on the Market1501 using metrics including rank-1, rank-5, rank-10, mAP, and percentages of annotated data. As illustrated in Table \ref{market}, AHSM utilizes only $43\%$ manual labeling to train an effective re-ID model. AHSM is slightly lower than random selection at rank-1 and mAP, \ie, about $97\%$ rank-1 and $95\%$ mAP of random selection, while AHSM can outperform the random selection at rank-5.

Besides, as illustrated in Figure \ref{fig5}(a), Figure \ref{fig5}(d), and Figure \ref{fig5}(g), the performance of AHSM is improved steadily with incrementally added training data, while the random selection performs unstably, which is mainly attributed to the fact that there are noisy images in the Market1501, such as low resolution, occlusion, and person detection errors. These images would introduce certain bias into the training set. On the contrary, AHSM yields smoother curve by reducing annotation of noisy images. Figure \ref{fig5}(a), Figure \ref{fig5}(d), and Figure \ref{fig5}(g) indicate that there is no obvious gap between our method and random selection in initialization, because query heuristics fail to elevate the performance quickly with small amount of queried images. But AHSM is able to provide a steady rise when more labeled images are added.

\textbf{Results on the MSMT17} In Table \ref{msmt}, we compared the performance of AHSM with random selection on the MSMT17 using metrics including rank-1, rank-5, rank-10, mAP, and percentages of annotated data. As illustrated in Table \ref{msmt}, AHSM utilizes only $37\%$ manual labeling to train an effective person re-ID model. AHSM falls behind the random selection at rank-1 and mAP, \ie, around $96\%$ rank-1 and $94\%$ mAP of random selection. At the same time, AHSM can outperform the random selection at rank-5.

Besides, as illustrated in Figure \ref{fig5}(b), Figure \ref{fig5}(e), and Figure \ref{fig5}(h), AHSM naturally yields steady and notable performance improvement with incrementally added training data, while random selection performs quite unstably. It should be noticed that AHSM outperforms the random selection at almost all the time. The main reason is the large scale dataset, since it can introduce high intra-class variation caused by large variations in appearance and illumination.

\begin{table}[!t]
	\setlength{\abovecaptionskip}{0.cm}
	\caption{Comparison on the Market1501 dataset. \protect\\ The best results are shown in boldface.}
	\label{market}
	\centering
	\renewcommand\arraystretch{1.5}
	\fontsize{8}{9}\selectfont
	\begin{tabular}{cccccc}
		\hline
		\multirow{2}*{Methods} & \multicolumn{5}{c}{Market1501} \\
		\cline{2-6}
		& rank-1 & rank-5 & rank-10 & mAP & Data(\%) \\ \hline
		EP~\cite{joshi2009multi} &0.8284&0.9361&0.9614&0.6479&85\% \\
		LC~\cite{scheffer2001active} &0.8022&0.9200&0.9500&0.6450&80\% \\
		MS~\cite{shannon1948mathematical} &0.8296&0.9073&0.9503&0.6252&50\%\\ \hline
		Random &0.8563&0.9368&0.9664&0.6825&100\% \\
		Our Method &\textbf{0.8391}&\textbf{0.9374}&\textbf{0.9623}&\textbf{0.6628}&\textbf{43\%} \\ \hline
 	\end{tabular}
\end{table}

\begin{table}[!t]
	\setlength{\abovecaptionskip}{0.cm}
	\caption{Comparison on the MSMT17 dataset. \protect\\ The best results are shown in boldface.}
	\label{msmt}
	\centering
	\renewcommand\arraystretch{1.5}
	\fontsize{8}{9}\selectfont
	\begin{tabular}{cccccc}
		\hline
		\multirow{2}*{Methods} & \multicolumn{5}{c}{MSMT17} \\
		\cline{2-6}
		& rank-1 & rank-5 & rank-10 & mAP & Data(\%) \\ \hline
		EP~\cite{joshi2009multi} & 0.5801 &0.7300&0.7877&0.3000&100\%\\
		LC~\cite{scheffer2001active} &0.5630&0.7116&0.7743&0.2812&90\%\\
		MS~\cite{shannon1948mathematical} &0.5867&0.7568&0.7908&0.2916&60\%\\ \hline
		Random &0.6048&0.6106&0.6652&0.3158&100\%\\
		Our Method &\textbf{0.5895}&0.7379&\textbf{0.7950}&\textbf{0.3058}&\textbf{37\%}\\ \hline
 	\end{tabular}
\end{table}

\begin{table}[!t]
	\setlength{\abovecaptionskip}{0.cm}
	\caption{Comparison on the CUHK03 dataset. \protect\\ The best results are shown in boldface.}
	\label{cuhk}
	\centering
	\renewcommand\arraystretch{1.5}
 	\fontsize{8}{9}\selectfont
	\begin{tabular}{cccccc}
		\hline
		\multirow{2}*{Methods} & \multicolumn{5}{c}{CUHK03} \\
		\cline{2-6}
		& rank-1 & rank-5 & rank-10 & mAP & Data(\%) \\ \hline
		EP~\cite{joshi2009multi} &0.8141&0.9630&0.9800&0.8444&80\%\\
		LC~\cite{scheffer2001active} &0.7958&0.9513&0.9767&0.8211&60\%\\
		MS~\cite{shannon1948mathematical} &0.8069&0.9493&0.9787&0.8193&70\% \\ \hline
		Random &0.8465&0.9715&0.9908&0.8743&100\% \\
		Our Method &\textbf{0.8264}&\textbf{0.9634}&\textbf{0.9878}&\textbf{0.8584}&\textbf{51\%}\\ \hline
 	\end{tabular}
\end{table}

\begin{table}[!t]
	\setlength{\abovecaptionskip}{0.cm}
	\caption{Comparison of Training Set size on the three datasets. \newline The best results are shown in boldface.}
	\label{size}
	\centering
	\renewcommand\arraystretch{1.5}
	\fontsize{8}{9}\selectfont
	\begin{tabular}{cccc}
		\hline
		\multirow{2}*{Methods} & \multicolumn{3}{c}{Dataset} \\
		\cline{2-4}
		& Market1501 & MSMT17 & CUHK03 \\ \hline
		EP~\cite{joshi2009multi} &10,996(85\%)&32,621(100\%)&8,327(80\%)\\
		LC~\cite{scheffer2001active} &12,349(80\%)&29,359(90\%)&6,245(60\%)\\
		MS~\cite{shannon1948mathematical} &6,468(50\%)&19,573(60\%)&7,286(70\%)\\ \hline
		Random &12,936(100\%)&32,621(100\%)&10,409(100\%) \\
		Our Method &\textbf{5,562(43\%)}&\textbf{12,070(37\%)}&\textbf{5,309(51\%)}\\ \hline
	\end{tabular}
\end{table}

\textbf{Results on the CUHK03}  In Table \ref{cuhk}, we compared the performance of AHSM with random selection on the CUHK03 using metrics including rank-1, rank-5, rank-10, mAP, and percentages of annotated data. As illustrated in Table \ref{cuhk}, AHSM utilizes only $51\%$ manual labeling samples to achieve almost the same performance at rank-1 and mAP as the random selection, \ie, around $96\%$ rank-1 and $97\%$ mAP of random selection. At the same time, AHSM outperforms random selection at rank-5.

Besides, as illustrated in Figure \ref{fig5}(c), Figure \ref{fig5}(f), and Figure \ref{fig5}(i), we exhibited the trade-off between manual labeling and accuracy to provide a steady rise when more labeled images are added. On the contrary, random selection yields quite unstable performance, which is mainly attributed to the small scale dataset with certain noisy samples. As a result, the elevation speed of AHSM slows from $0\%$ to $25\%$, but can steadily increase with incrementally added training data.

From the results shown in Table \ref{market}, \ref{msmt} and \ref{cuhk}, we can conclude that the labeling efforts can be significantly reduced by hard sample mining and redundancy reduction. It is worth noticing that AHSM gives the best performance on the MSMT17 dataset, the largest public person re-ID thus far.

\subsubsection{\textbf{Comparison with Query Heuristics}}
In Table \ref{market}, \ref{msmt}, and \ref{cuhk}, we compared the performance of AHSM with three query heuristics on the three datasets at different ranks and mAP, where the comparison curves for performance improvement are listed in Figure \ref{for4}.

To verify the effectiveness of AHSM, we compared it to EP, LC, and MS on the three datasets. With EP, LC, and MS, we ranked the samples by their scores and selected the top-ranked samples for annotation. When comparing AHSM with them, AHSM presents a huge advantage in both keeping the performance of the re-ID model and reducing the labeling efforts. As illustrated in Table \ref{market}, the query heuristics almost fail to reduce the labeling efforts for the re-ID model. We can observe the large efforts reduction gap between AHSM and query heuristic. As shown in Figure \ref{msmt}, \ref{cuhk} and Figure \ref{fig5}, the EP heuristic fails to accelerate the convergence of the re-ID model on the three datasets. It should be noticed that the EP heuristic performs even worse than random selection. We can also observe that LC and MS may make the re-ID model converge earlier than AHSM but with the final inferior performance. Thus, AHSM would get a better trade-off between performance and labeling efforts.

As shown in Table \ref{market}, \ref{msmt} and \ref{cuhk}, we can conclude that AHSM performs the best compared to the commonly used query heuristics. At the same time, AHSM significantly reduces labeling efforts via hard sample mining and redundancy reduction.

\subsection{Ablation Study for AHSM}
We conducted comprehensive ablation studies on the three datasets to explore the function of uncertainty and intra-diversity estimation. The results at the metrics: rank-1, rank-5, rank-10, and mAP are shown in Table \ref{uncertain} and \ref{diversity} and Figure \ref{fig6}. Each result is obtained with only one estimation and the rest settings are the same as the default.

\begin{table}[!t]
	\setlength{\abovecaptionskip}{0.cm}
	\caption{Performance of Uncertainty Estimation.}
	\label{uncertain}
	\centering
	\renewcommand\arraystretch{1.5}
	\fontsize{8}{9}\selectfont
	\begin{tabular}{cccccc}
		\hline
		\textbf{ }  & \textbf{Method}   &rank-1 & rank-5 & rank-10 & mAP  \\ \hline
		\multirow{2}*{Market1501} &\textbf{AHSM}  &0.8328&0.9374&0.9623&0.6528\\
		~&\textbf{\scriptsize{Uncertainty}} &0.8130&0.9121&0.9386&0.6353\\\hline
		\multirow{2}*{MSMT17} &\textbf{AHSM}  &0.5805&0.7379&0.7950&0.2963\\
		~&\textbf{\scriptsize{Uncertainty}}  &0.5388&0.7084&0.7700&0.2700\\\hline
		\multirow{2}*{CUHK03} &\textbf{AHSM}&0.8161&0.9634&0.9878&0.8484 \\
		~&\textbf{\scriptsize{Uncertainty}}  &0.7967&0.9460&0.9600&0.8232\\ \hline
	\end{tabular}
\end{table}
\begin{table}[!t]
	\setlength{\abovecaptionskip}{0.cm}
	\caption{Performance of Intra-diversity Estimation.}
	\label{diversity}
	\centering
	\renewcommand\arraystretch{1.5}
	\fontsize{8}{9}\selectfont
	\begin{tabular}{p{10mm}ccccc}
		\hline
		\textbf{ }  & \textbf{Method}   &rank-1 & rank-5 & rank-10 & mAP  \\ \hline
		\multirow{2}*{Market1501} &\textbf{AHSM}  &0.8328&0.9374&0.9623&0.6528\\
		~&\textbf{\scriptsize{Intra-diversity}} &0.8195&0.9284&0.9501&0.6400\\\hline
		\multirow{2}*{MSMT17} &\textbf{AHSM}  &0.5805&0.7379&0.7950&0.2963\\
		~&\textbf{\scriptsize{Intra-diversity}}  &0.5563&0.7150&0.7705&0.2522\\\hline
		\multirow{2}*{CUHK03} &\textbf{AHSM}&0.8161&0.9634&0.9878&0.8484 \\
		~&\textbf{\scriptsize{Intra-diversity}}  &0.7945&0.9481&0.9600&0.8271\\ \hline
	\end{tabular}
\end{table}

\subsubsection{\textbf{Uncertainty Estimation Analysis}}
First, we merely adopted uncertainty estimation to select samples to annotate. As illustrated in Table \ref{uncertain}, the performance of uncertainty estimation is lower than that of AHSM at all metrics. As shown in Figure \ref{fig6}, the green/triangle curves denote the performance improvement of uncertainty estimation can converge quicker, but with a low recognition accuracy on the three datasets. The reason may lie in that the model converges to a local minimum. We can conclude that: (I) Uncertainty estimation only exhibits weaker accuracy than that of the AHSM. (II) The uncertainty estimation can make the re-ID model converge faster. And (III) the uncertainty estimation may make the model converge to a local minimum. It clearly verifies the uncertainty estimation can influence the convergence of the model.

\subsubsection{\textbf{Intra-diversity Estimation Analysis}}
Second, we merely employed intra-diversity estimation to select samples to annotate. As illustrated in Table \ref{diversity}, the performance of uncertainty estimation is weaker than that of AHSM at all metrics. As shown in Figure \ref{fig6}, the blue/plus curves denote the convergence speed of intra-diversity estimation is slower compared to AHSM and uncertainty, but with a slightly higher recognition accuracy compared with uncertainty. From Table \ref{uncertain} and Figure \ref{fig6}, we can obtain: (I) Intra-diversity estimation delivers slightly weaker accuracy than that of the AHSM, but outperforms uncertainty. And (II) the intra-diversity estimation exhibits a slow convergence speed of the model. It shows the intra-diversity estimation is able to influence the performance of the model.

From the results of ablation studies shown in Table \ref{uncertain} and \ref{diversity}, we can conclude that the uncertainty estimation can select effective hard samples and contribute to converge the model. Besides, the intra-diversity estimation can further improve the performance of the re-ID model via reducing the redundant hard samples.
\begin{figure*}[!t]	
	\setlength{\abovecaptionskip}{0.cm}
	\setlength{\belowcaptionskip}{-0.cm}
	\centering
	\includegraphics[width=1\linewidth]{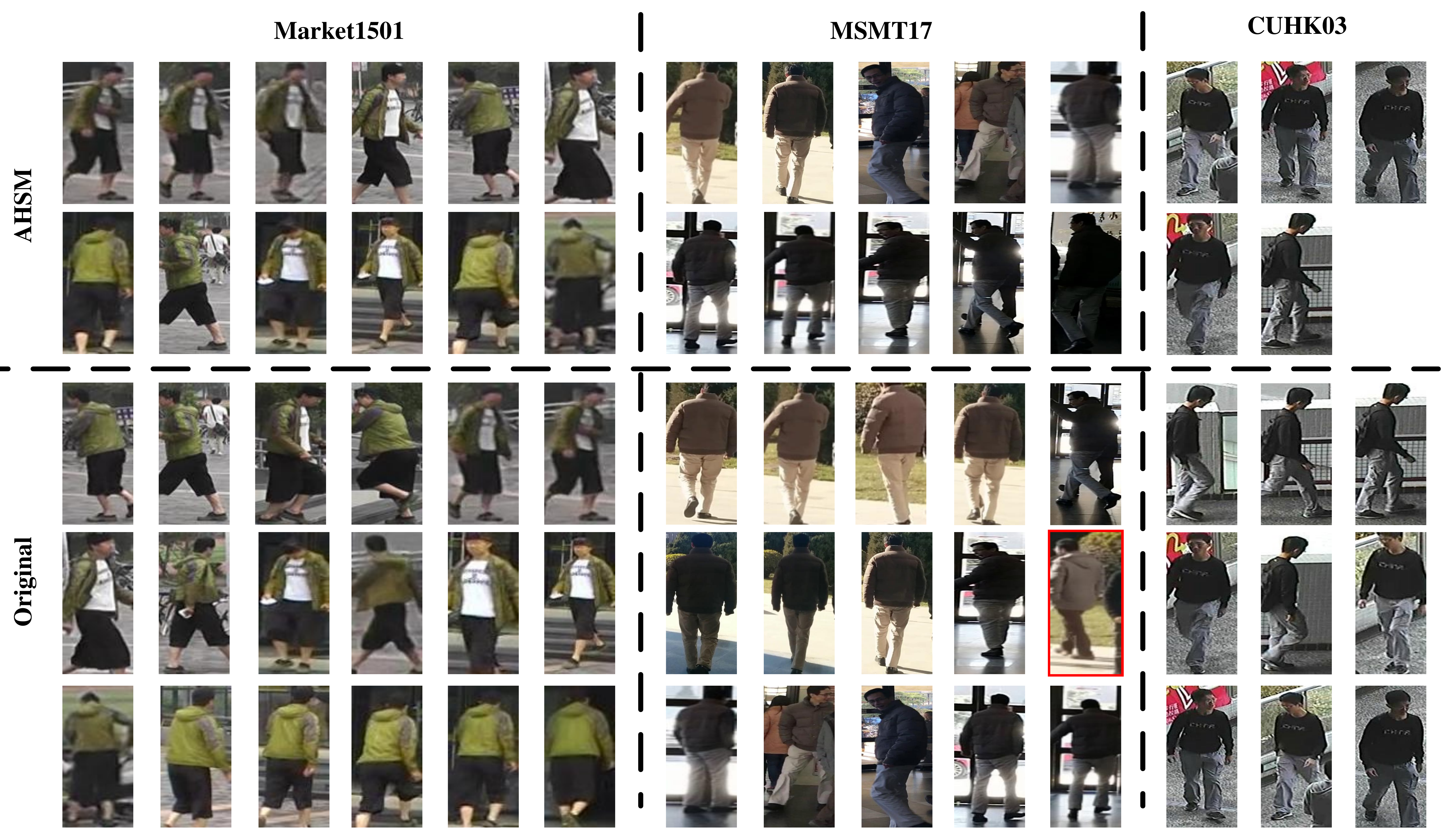}
	\caption{Effective hard samples selection via AHSM vs. the original samples of a given person on the three datasets. Red rectangle is a wrongly labeled sample on the MSMT17 dataset.}
	\label{fig7}
\end{figure*}

\subsection{Hard Sample Mining Analysis}
Hard sample mining is embedded into the active leaning framework via uncertainty estimation and intra-diversity estimation. AHSM is able to successfully train an effective re-ID model with the least labeling efforts as shown in experimental results. Incorporating with hard sample mining, active leaning scheme can select hard samples with more informative patterns to reduce labeling efforts.

As shown in Figure \ref{fig5}, Table \ref{uncertain}, and Table \ref{diversity}, the ablation studies demonstrated the effectiveness of the uncertainty and intra-diversity estimation. Uncertainty estimation can select the hard samples to accelerate the convergence of the re-ID model, while intra-diversity estimation can achieve higher performance than uncertainty estimation. Working with these two estimation methods, AHSM is able to make a trade-off between performance and labeling efforts. Besides, AHSM exhibits competitive results both in performance and labeling efforts compared with other active learning heuristics.

\begin{table}[!t]
	\setlength{\abovecaptionskip}{0.cm}
	\caption{Comparison of Labeling Efforts on the Market1501.}
	\label{mtime}
	\centering
	\renewcommand\arraystretch{1.5}
	\fontsize{8}{9}\selectfont
	\setlength{\tabcolsep}{5mm}{
	\begin{tabular}{ccc}
		\hline
		\textbf{Dataset}  & \textbf{Method}   &Comparison Times \\ \hline
		
		\multirow{10}*{Market1501} &  \textbf{AHSM}  &$7.7\times10^{5}$(43\%)\\
		~  &  \textbf{AHSM+IDRM}  &$1.2\times10^{4}$(43\%)\\ \cline{2-3}
		~  &  \textbf{EN}  &$1.5\times10^{6}$(85\%)\\
		~  &  \textbf{EN+IDRM}  &$2.7\times10^{4}$(85\%)\\ \cline{2-3}
		~  &  \textbf{LC}  &$1.4\times10^{6}$(80\%)\\
		~  &  \textbf{LC+IDRM}  &$2.5\times10^{4}$(80\%)\\ \cline{2-3}
		~  &  \textbf{MS}  &$9.0\times10^{5}$(50\%)\\
		~  &  \textbf{MS+IDRM}  &$1.3\times10^{4}$(50\%)\\ \cline{2-3}
		~  &  \textbf{Random}  &$1.8\times10^{6}$(100\%)\\	
		~  &  \textbf{Random+IDRM}  &$2.6\times10^{5}$(100\%)\\ \hline
		
	\end{tabular}}
\end{table}

\begin{table}[!t]
	\setlength{\abovecaptionskip}{0.cm}
	\caption{Comparison of Labeling Efforts on the MSMT17.}
	\label{mstime}
	\centering
	\renewcommand\arraystretch{1.5}
	\fontsize{8}{9}\selectfont
	\setlength{\tabcolsep}{5mm}{
	\begin{tabular}{ccc}
		\hline
		\textbf{Dataset}  & \textbf{Method}   &Comparison Times \\ \hline
		
		\multirow{10}*{MSMT17}  &  \textbf{AHSM}  &$3.0\times10^{6}$(37\%)\\
		~  &  \textbf{AHSM+IDRM}&$2.6\times10^{5}$(37\%) \\ \cline{2-3}
	    ~  &  \textbf{EN}  &$8.0\times10^{6}$(100\%)\\
		~  &  \textbf{EN+IDRM}  &$4.9\times10^{5}$(100\%)\\ \cline{2-3}
		~  &  \textbf{LC}  &$7.2\times10^{6}$(90\%)\\
		~  &  \textbf{LC+IDRM}  &$4.9\times10^{5}$(90\%)\\ \cline{2-3}
		~  &  \textbf{MS}  &$4.8\times10^{6}$(60\%)\\
		~  &  \textbf{MS+IDRM}  &$4.2\times10^{5}$(60\%)\\ \cline{2-3}
		~  &  \textbf{Random} &$8.0\times10^{6}$(100\%)\\
		~  &  \textbf{Random+IDRM}  &$6.5\times10^{5}$(100\%)\\ \hline
		
	\end{tabular}}
\end{table}

\begin{table}[!t]
	\setlength{\abovecaptionskip}{0.cm}
	\caption{Comparison of Labeling Efforts on the MSMT17.}
	\label{ctime}
	\centering
	\renewcommand\arraystretch{1.5}
	\fontsize{8}{9}\selectfont
	\setlength{\tabcolsep}{5mm}{
	\begin{tabular}{ccc}
		\hline
		\textbf{Dataset}  & \textbf{Method}   &Comparison Times \\ \hline
	   \multirow{10}*{CUHK03}  &  \textbf{AHSM}  &$2.1\times10^{6}$(51\%)\\
		~  &  \textbf{AHSM+IDRM}  &$9.6\times10^{4}$(51\%) \\ \cline{2-3}
	    ~  &  \textbf{EN}  &$3.3\times10^{6}$(80\%)\\
		~  &  \textbf{EN+IDRM}  &$2.1\times10^{5}$(80\%)\\ \cline{2-3}
		~  &  \textbf{LC}  &$2.5\times10^{6}$(60\%)\\
		~  &  \textbf{LC+IDRM}  &$1.6\times10^{5}$(60\%)\\ \cline{2-3}
		~  &  \textbf{MS}  &$2.9\times10^{6}$(70\%)\\
		~  &  \textbf{MS+IDRM}  &$1.8\times10^{5}$(70\%)\\ \cline{2-3}
		~  &  \textbf{Random}  &$4.1\times10^{6}$(100\%)\\
	    ~  &  \textbf{Random+IDRM}  &$2.6\times10^{5}$(100\%)\\ \hline
	\end{tabular}}
\end{table}

\subsection{Labeling Efforts Analysis}
We recorded the comparison times of IDRM and random selection for every $10\%$ data to reflect the labeling efforts. The comparison times are highly correlated to the amount of identities in the dataset. As illustrated in Table \ref{mtime}, \ref{mstime}, and \ref{ctime}, the comparison times of IDRM are about $1\%$, $10\%$, and $1\%$ of the random selection in the Market1501, MSMT17, and CUHK03, respectively. It should be noticed the human annotators need to compare nearly millions of times for these datasets, \ie, it typically involves thousands of hours of human efforts.

Figure \ref{fig7} illustrates the selected hard samples via AHSM and the original samples of a given person. The selected images via AHSM contain comprehensive information of the original samples. Besides, it should be noticed that AHSM can avoid wrongly annotated images with the help of IDRM and human annotators. For example, the image with a red rectangle in Figure \ref{fig7} is wrongly labeled in the original samples, and does not appear in the selected images via AHSM. The aforementioned tables and figures demonstrate that the labeling efforts can be significantly reduced by the AHSM and IDRM.

\section{Conclusion and Future Work}
In this work, we present an active learning framework to address the data collection problem in the task of person re-ID. To select hard samples for training, the uncertainty estimation is designed to evaluate the prediction confidence of the current model for a sample, and intra-diversity estimation is designed to optimize the intra-class sample distribution. To further reduce the labeling efforts, a computer-assisted interface IDRM is developed to help human annotators to accurately label the selected samples efficiently. Experiments on the Market1501, MSMT17 and CUHK03 datasets have demonstrated that AHSM can minimize annotation efforts while maximizing the performance of the person re-ID model.

In future, we plan to deploy our AHSM framework to facilitate the construction of large-scale re-ID dataset, where active hard sample mining needs to be addressed.





\ifCLASSOPTIONcaptionsoff
  \newpage
\fi



\bibliographystyle{IEEEtran}
\bibliography{reference}
\end{document}